\documentclass[11pt,a4paper]{article}
\usepackage[hyperref]{acl2021}
\usepackage{times}
\usepackage{latexsym}
\usepackage{siunitx}
\usepackage{graphicx}
\usepackage{xcolor}

\usepackage{latexsym}
\usepackage{siunitx}
\usepackage{graphicx}
\usepackage{xcolor}
\usepackage{float}
\usepackage{placeins}
\usepackage{amssymb}
\usepackage{pifont}
\newcommand{\xmark}{\ding{55}}%
\graphicspath{ {./graphics/} }

\graphicspath{ {./graphics/} }

\usepackage{microtype}

\usepackage[inline]{enumitem}
\usepackage{pmboxdraw}

\aclfinalcopy 
\def\aclpaperid{4016} 

\title{Are Multilingual Models the Best Choice for Moderately Under-Resourced Languages? A Comprehensive Assessment for Catalan}

\author{Jordi Armengol-Estap\'{e}, {\bf Casimiro Pio Carrino}, {\bf Carlos Rodriguez-Penagos,} \\  {\bf Ona de Gibert Bonet,} {\bf Carme Armentano-Oller,} {\bf Aitor Gonzalez-Agirre,} \\ {\bf Maite Melero,} {\bf Marta Villegas}\\
Barcelona Supercomputing Center, Barcelona, Spain\\
    \texttt{\{jordi.armengol, casimiro.carrino,  carlos.rodriguez1,}\\\texttt{ ona.degibert, carme.armentano, aitor.gonzalez,}\\ \texttt{maite.melero, marta.villegas\}@bsc.es}\\}

\date{}

\begin{document}
\maketitle
\begin{abstract}

Multilingual language models have been a crucial breakthrough as they considerably reduce the need of data for under-resourced languages. Nevertheless, the superiority of language-specific models has already been proven for languages having access to large amounts of data.
In this work, we focus on Catalan with the aim to explore to what extent a medium-sized monolingual language model is competitive with state-of-the-art large multilingual models. For this, we: (1) build a clean, high-quality textual Catalan corpus (CaText), the largest to date (but only a fraction of the usual size of the previous work in monolingual language models), (2) train a Transformer-based language model for Catalan (BERTa), and (3) devise a thorough evaluation in a diversity of settings, comprising a complete array of downstream tasks, namely, Part of Speech Tagging, Named Entity Recognition and Classification, Text Classification, Question Answering, and Semantic Textual Similarity, with most of the corresponding datasets being created ex novo. The result is a new benchmark, the Catalan Language Understanding Benchmark (CLUB), which we publish as an open resource, together with the clean textual corpus, the language model, and the cleaning pipeline. Using state-of-the-art multilingual models and a monolingual model trained only on Wikipedia as baselines, we consistently observe the superiority of our model across tasks and settings.

\end{abstract}

\section{Introduction}
Over the past decades, Natural Language Processing (NLP) has become a powerful technology that may be found behind many AI-based consumer products, such as voice assistants, automatic translators, intelligent chatbots, etc. This undeniable success is somehow tarnished by the fact that most NLP resources and systems are available only for a small percentage of languages \cite{joshi2021state}.

In contrast, most of the languages spoken in the world today, even some with millions of speakers, are left behind, both in the research and in the development of the technology.
Recent breakthroughs in deep learning, specifically the Transformer architecture \cite{DBLP:journals/corr/VaswaniSPUJGKP17}, have revolutionized the entire field and have opened the doors to powerful transfer learning and unsupervised techniques, making it possible for under-resourced languages to benefit -at least, partially- from the formidable advances taking place for English.
Transformed-based multilingual pre-trained models \cite{DBLP:journals/corr/abs-1810-04805, DBLP:journals/corr/abs-1911-02116} soon showed an impressive increase of performance also for  under-resourced languages, as they considerably reduce the amount of training data needed for a particular task.

The question as to whether training  language-specific models was worth the effort, given those impressive results, was more or less quickly resolved, for languages having enough monolingual data to train with \cite{DBLP:journals/corr/abs-1911-03894, DBLP:journals/corr/abs-1912-07076}. However, for most languages it is still a challenge to obtain such large amounts of data. Therefore, the question still stands for many of them.
There is a huge variation in the potential access to language resources for any given language, going from high-resourced, medium, under or severely under-resourced.

In this paper, we focus on Catalan, a moderately under-resourced language. By comparison, the size of the corpus -purposefully collected and cleaned for this work- is almost half of the one in the Finnish FinBERT \cite{DBLP:journals/corr/abs-1912-07076} (a comparably sized language), and almost 20 times smaller than the French CamemBERT. Another defining characteristic of Catalan is its affiliation to the Romance family, which is abundantly represented in multilingual pre-trained models. Several big languages (French, Spanish, Italian, Portuguese) belong to this family and, thus, are typologically close to Catalan. Presumably, this fact could give a Romance-rich multilingual model an upper hand in its comparison with a Catalan-only model.

This exercise also gives us the opportunity to enrich the number and quality of open-source resources for Catalan. Our contributions can be summarized as follows:
\begin{itemize}
    \item We compile a clean, high-quality textual Catalan corpus, the largest to date, released with an open license.
    \item We build a complete end-to-end cleaning pipeline, released as open-source. 
    \item We train a Transformer-based language model for Catalan, also openly released.
    \item We create new annotated corpora for those tasks for which there was not any, such as Text Classification (TC), Question Answering (QA) and Semantic Textual Similarity (STS). We publicly release them as well. Together with existing Part-Of-Speech (POS) and Named Entity Recognition and Recognition (NERC) datasets, they are part of a new Natural Language Understanding (NLU) benchmark, the Catalan Language Understanding Benchmark (CLUB).
\end{itemize}

All the code, datasets and the final model are made available to the community in standard formats.\footnote{\url{https://github.com/TeMU-BSC/berta}}

\section{Previous Work}
While the original motivation for the Transformer was machine translation, it has been successfully applied to a wide range of tasks, excelling at representation learning via unsupervised pre-training. Both in decoder-based language models, as in GPT \cite{Radford2018ImprovingLU}, and in encoder-based masked language models, as pioneered by BERT \cite{DBLP:journals/corr/abs-1810-04805}, large Transformer models are pre-trained on big unlabelled corpora. The learned representations can then be applied as a feature extractor or by fine tuning to the downstream task of choice, typically resulting in state-of-the-art performance.

The original BERT also had a multilingual version, mBERT, leveraging monolingual text from the corresponding Wikipedia of different languages. In XLM \cite{DBLP:journals/corr/abs-1901-07291}, authors introduced a \textit{cross-lingual} language model, explicitly modeling cross-lingual representations (instead of just concatenating text from all languages). This model was scaled up producing XLM-RoBERTa \cite{DBLP:journals/corr/abs-1911-02116}, based on RoBERTa \cite{DBLP:journals/corr/abs-1907-11692}, a variant of BERT with a simplified pre-training objective.

Especially in the case of BERT-like models, which are intended for NLU tasks (rather than generation), the literature has been considerably prolific in terms of language-specific models. While both mBERT and XLM-RoBERTa are considered the state-of-the-art in NLU for numerous languages, several works observed that language-specific models trained from scratch proved to obtain better performance, such as the French CamemBERT \cite{DBLP:journals/corr/abs-1911-03894}, the Dutch BERTje \cite{DBLP:journals/corr/abs-1912-09582}, and the Finnish FinBERT \cite{DBLP:journals/corr/abs-1912-07076}. FinBERT authors made emphasis on text cleaning, claimed to be essential. The same authors proposed WikiBERT \cite{pyysalo2020wikibert}, a set of language-specific baselines based on BERT for as many as 42 languages (including Catalan). Regarding low-resource languages, Basque-specific models have been shown to outperform mBERT \cite{agerri2020text}, although it is worth pointing out that Basque, being a linguistic  isolate, is typologically far apart from the rest of languages in the pre-training corpus of mBERT.


The authors of these language-specific models hypothesized different causes behind the increase in performance with respect to the multilingual models: \begin{enumerate}
    \item Having a language-specific vocabulary, they avoid the split of words into too many subwords (which, linguistically, are less interpretable);
    \item The amount of language-specific data; and
    \item Training on more diverse data of the target language (e.g., web crawlings, instead of just Wikipedia)
\end{enumerate}. In \citet{nozza2020mask}, they compare the performance of mBERT with a number of language-specific models. They conclude that there is a huge variability in the models and that it is difficult to find the best model for a given task, language, and domain.



In this work, we further investigate the issue of whether building language-specific models from scratch is worth the effort, and if so under which circumstances. Unlike previous works, we focus on a moderately under-resourced language, Catalan. In addition to being low-resource, Catalan is close to other Romance languages largely present in the pre-training corpora of multilingual models, and knowledge from these other Romance languages transfers well to Catalan. These two facts call into question whether building language-specific models for these cases is still interesting. We show that a Catalan-specific model is indeed relevant, and we provide tools and recipes for other languages in a similar situation. Besides the model itself, we build the largest Catalan (clean) pre-training corpus to date as well as an extensive evaluation benchmark for NLP tasks in Catalan.


\section{Pre-training Corpus}
\subsection{Data sources}
\label{sec:datasources}

Our new Catalan text corpus, CaText, includes both data from datasets already available in Catalan and data from three new crawlers we recently ran.

From the published datasets, we use (1) the Catalan part of the DOGC corpus, a set of documents from the Official Gazette of the Catalan Government; (2) the Catalan Open Subtitles, a collection of translated movie subtitles \cite{tiedemann2012parallel}; (3) the non-shuffled version of the Catalan part of the OSCAR corpus \cite{suarez2019asynchronous}, a collection of monolingual corpora, filtered from Common Crawl;\footnote{\url{https://commoncrawl.org/about/}} (4) the CaWac corpus, a web corpus of Catalan built from the \texttt{.cat} top-level-domain in late 2013 \cite{ljubesic2014cawac}, the non-deduplicated version, and (5) the Catalan Wikipedia articles downloaded on 18-08-2020.

Regarding the newly created datasets, we ran three new crawlings: (6) the Catalan General Crawling, obtained by crawling the 500 most popular \texttt{.cat} and \texttt{.ad} domains; (7) the Catalan Government Crawling, obtained by crawling the \url{gencat.cat} domain and subdomains, belonging to the Catalan Government; and (8) the ACN corpus with 220k news items from March 2015 until October 2020, crawled from the Catalan News Agency.\footnote{\url{https://www.acn.cat/}}

\subsection{Preprocessing}

In order to be able to obtain a high-quality training corpus, we apply strict filters to the raw data, by means of a cleaning pipeline built for the purpose, CorpusCleaner.\footnote{\url{https://github.com/TeMU-BSC/corpus-cleaner-acl}} This pipeline supports 100+ languages\footnote{We take this number from the used language identifiers. We have actually tested the cleaning pipeline with a variety of languages and domains, such as Basque, Finnish, Kazakh, Georgian and Biomedical Spanish, among others.} and stands out for keeping document boundaries, instead of being sentence-based, which allows modeling long-range dependencies.

\paragraph{Transforms:} CorpusCleaner is able to parse data in different formats (WARC files from crawlings, for instance), in a way that document boundaries and metadata are kept whenever possible. It includes the \texttt{ftfy} library \cite{speer-2019-ftfy} for fixing encoding errors and applies raw string transformations for normalizing spaces, removing HTML tags, and others.

\paragraph{Filters:} The pipeline applies a \textit{cascade of language identifiers} (similarly to \citealp{8345541}), using FastText's \cite{bojanowski2016enriching} language identifier and LangId.\footnote{\url{https://github.com/saffsd/langid.py}} By cascade, we mean that we first apply the faster language identifiers to discard documents in which we are sure that the target language is not present. We then use the slower (but with better performance) language identifiers only with the remaining documents. After that, we make use of a fast and tokenization-agnostic sentence splitter.\footnote{\url{https://github.com/mediacloud/sentence-splitter}} In addition, CorpusCleaner includes numerous heuristics (e.g., standard deviation of sentence length in a given document for detecting badly split sentences coming from PDFs, or non-natural text) that apply reasonably well to the languages we have tested so far. Some of these rules have been inspired by \citet{DBLP:journals/corr/abs-1912-07076}. Even in the case of document-level corpora, some rules are also applied at sentence-level, to improve document coherency (e.g., remove corrupted sentences in an otherwise high-quality document). 

\paragraph{Deduplication:} The last stage of the pipeline consists of document-level deduplication, based on n-gram repetitions with Onion \cite{11858/00-097C-0000-000D-F67B-7}. In addition, we also deduplicate at the sentence-level with a threshold of occurrences, since most of the often repeated sentences are not natural linguistic occurrences, but placeholders commonly used by web developers (e.g. copyright notices). 





Table \ref{tab:corpus_datasets} shows the composition of the CaText corpus. 

\begin{table}
\centering
\begin{tabular}{lrr}
\hline
\textbf{ } \textbf{Dataset} & \textbf{Original} & \textbf{Final} \\
\hline
1 DOGC & 126.65 &  126.65 \\
2 Cat. Open Subtitles & 3.52 & 3.52 \\
3 Cat. OSCAR & 1,355.53 & 695.37 \\
4 CaWaC & 1,535.10 & 650.98 \\
5 Wikipedia & 198.36& 167.47 \\
6 Cat. Gen. Crawling & 1,092.98 & 434.82 \\
7 Cat. Gov. Crawling & 303.10 & 39.12 \\
8 ACN & 81.28 & 75.61 \\
\hline
Total& 4,696.52 & 2,193.54 \\
Deduplicated (CaText) &  & 1,770.32\\
\hline
\end{tabular}
\caption{\label{tab:corpus_datasets} Number of tokens (in millions) in the different used corpora before and after the filtering process, just before deduplication. The difference between these two columns shows the scale of the clean-up performed by the filters in the pipeline, particularly in the crawled data: OSCAR, CaWaC, Catalan General Crawling and Catalan Government Crawling. The last row the number of tokens after a global document deduplication across all corpora.}
\end{table}

\paragraph{Splits and release:} We sample 2,000 documents for validation, in order to monitor the training, and 2,000 more for test, as a hold-out set for future analysis. We release the preprocessed corpus CaText\footnote{\url{http://doi.org/10.5281/zenodo.4636228}} with an open license, but respecting the licenses of the original data sources.\footnote{Regarding the two crawlings run in-house, we have also opted to release them with a more open license so they can be used freely (see \url{https://zenodo.org/record/4636228} and \url{https://zenodo.org/record/4636899}).} 

\subsection{Vocabulary}

To build the language model, we use Byte-Level BPE \cite{radford2019language}, as in the original RoBERTa, but learning the vocabulary from scratch using the training set of CaText. Following recent works, we keep casing, and use a vocabulary size of 52k tokens \cite{scheible2020gottbert}.

As shown in Table \ref{tab:tokenizations}, our tokenization, being language-specific (and having a bigger vocabulary than WikiBERT-ca), generates less subwords per word, which has been shown to be beneficial \cite{DBLP:journals/corr/abs-1912-07076, agerri2020text}.

We describe the models used in our comparative study in Section \ref{sec:modelcat} and Section \ref{sec:evaluation}. Here, we provide some examples of the different tokenization results depending on the used tokenizer:


\begin{small}
\begin{verbatim}

original: coronavirus
BERTa: coronavirus
mBERT: corona ##vir ##us
WikiBERT-ca: corona ##vir ##us
XLM-RoBERTa: ▁corona virus

original: lamentablement
BERTa: lamentablement
mBERT: la ##menta ##blement
WikiBERT-ca: la ##menta ##ble ##ment
XLM-RoBERTa: ▁lamentable ment

\end{verbatim}
\end{small}

\begin{table}
\centering
    \begin{tabular}{lr}
    \hline
    \textbf{Model} & \textbf{Subwords per sentence} \\
    \hline
    BERTa & \textbf{33.94} \\
    mBERT & 41.14 \\
    WikiBERT-ca & 38.38\\
    XLM-RoBERTa & 38.62 \\
\hline
\end{tabular}
\caption{\label{tab:tokenizations} Subwords per sentence in the test set of CaText.}
\end{table}

As the examples show, our Catalan-specific model (BERTa) is more likely to keep the full word, even in the presence of derivative morphemes, such as the adverb 'lamentablement' (\textit{unfortunately}). The fact that it does not split the word 'coronavirus' either reveals that a large part of the training corpus is very recent and includes many references to the COVID-19 pandemic. In the Appendix \ref{app:vocab} we provide more detailed information about vocabulary overlapping between models.


\section{CLUB: The Catalan Language Understanding Benchmark}

In order to evaluate our model on different downstream tasks, we generate a new benchmark for evaluating NLU capabilities for Catalan, the Catalan Language Understanding Benchmark (CLUB).\footnote{See \url{https://github.com/TeMU-BSC/berta}.} To build it, we bootstrap from existing resources (such as the Ancora corpus\footnote{\url{https://doi.org/10.5281/zenodo.4762030}}) and create new high quality ones from scratch, adopting (and in some cases improving on) existing guidelines for well known benchmarks. These datasets are publicly available in the Zenodo platform,\footnote{\url{https://zenodo.org/}} under the Catalan AI language resources community, where further details about the curation and annotation guidelines for each one are given. In general, we provide as much information as possible following \citet{10.1162/tacl_a_00041} guidelines when relevant. For example, gender and socioeconomic status are considered not as relevant for the kind of semantic annotations created. However, the fact that all commissioned annotators (1) were native speakers of Catalan, (2) were translators, editors, philologists or linguists, and (3) had previous experience in language-related tasks, \textit{is} considered to be important. The curation rationale we follow wis to make these datasets both representative of contemporary Catalan language use, as well as directly comparable to similar reference datasets from the General Language Understanding Evaluation (GLUE) benchmark \cite{DBLP:journals/corr/abs-1804-07461}.\footnote{\url{https://gluebenchmark.com/}}  

For \textbf{Part-of-Speech Tagging (POS) }and \textbf{Named Entity Recognition and Classification (NERC),} we use annotations from the Universal Dependencies treebank\footnote{\url{https://github.com/UniversalDependencies/UD_Catalan-AnCora}} of the well-known Ancora corpus, which uses UPOS tags. We extracted named entities from the original Ancora\footnote{\url{https://doi.org/10.5281/zenodo.4762030}} version, filtering out some unconventional ones, like book titles, and transcribe them into a standard CONLL-IOB format.

For \textbf{Semantic Textual Similarity (STS)} \cite{agirre-etal-2012-semeval}, we create a new dataset from scratch. 
We use different similarity measures (Jaccard, Doc2Vec \cite{DBLP:journals/corr/LeM14} and DistilBERT \cite{DBLP:journals/corr/abs-1910-01108} embedding cosine similarity) to pre-select potential sentence pairs from the aforementioned CaText corpus, as well as a final manual review to ensure that the selection represented superficial and deep similarities in subject matter and lexicon. This results in 3,073 pairs for manual annotation. Following the guidelines set in the series of landmark SemEval challenges,\footnote{\url{http://ixa2.si.ehu.eus/stswiki}} we commission 4 native speaker annotators to assess the similarity of the sentence pairs on a scale between 0 (\textit{completely dissimilar}) to 5 (\textit{completely equivalent}), with other possible values, such as 3 (\textit{roughly equivalent, but some important information differs}). Then, for each sentence pair, we compute the mean of the four annotations, and we discard single annotations that deviate by more than 1 from the mean. After this cleaning process, we use the mean of the remaining annotations as a final score. Finally, in order to assess the annotation quality of the dataset, we measure the correlation of each annotator with the average of the rest of the annotators, and average all the individual correlations, resulting in a Pearson correlation of 0.739.\footnote{\url{https://doi.org/10.5281/zenodo.4529183}}

For \textbf{Text Classification (TC)}, we use 137k news pieces from the Catalan News Agency (ACN) corpus, mentioned in Section \ref{sec:datasources}. As labels in our classification task, we used the article category provided by the metadata, keeping only those categories that had more than 2,000 articles. See the Appendix \ref{app:datasets} for more details on the distribution by label. We call this benchmark TeCla (Text Classification Catalan dataset).\footnote{\url{https://doi.org/10.5281/zenodo.4627197}}

Finally, for extractive \textbf{Question Answering (QA)}, we compile two datasets:
\begin{itemize}
    \item The Catalan translation of XQuAD \cite{artetxe2020crosslingual}, a multilingual collection of manual translations of fragments from English Wikipedia articles used mainly for cross-lingual analyses. The Catalan dataset, XQUAD-ca,\footnote{\url{https://doi.org/10.5281/zenodo.4526223}} as the rest of languages, includes a subset of 240 paragraphs and 1,190 question-answer pairs from the development set of SQuAD v1.1 \cite{rajpurkar2016squad}, and has no adversarial or unanswerable questions.
    \item A new dataset, ViquiQuAD, an extractive QA dataset from Catalan Wikipedia)\footnote{\url{https://doi.org/10.5281/zenodo.4562344}} consisting of more than 15,000 questions outsourced from Catalan Wikipedia. We randomly choose a set of 596 articles which were originally written in Catalan, i.e. not translated from other Wikipedias. From those, we randomly select 3,129 short paragraphs to use as contexts and ask annotators to create up to 5 questions that could be answered by quoting directly from the context provided. In the Appendix \ref{app:datasets}, we provide some statistics on our QA datasets and list the types of questions, comparing ViquiQuAD and XQuAD-ca with XQuAD-en and the French FQuAD \cite{dhoffschmidt2020fquad}. 
\end{itemize}

These benchmarks present a well-balanced set of challenges with which to test and compare our model with others. Apart from the diversity in the tasks themselves, for QA we provide an additional test from a different distribution (XQuAD-ca), and the data regime is heterogeneous across datasets (from more than 100k samples in TC to 3k in STS).

\begin{table}
\centering
\begin{tabular}{lrrrr}
\hline \textbf{Task} & \textbf{Total} & \textbf{Train} & \textbf{Dev} & \textbf{Test} \\ 
\hline
NERC	& 13,581	& 10,628	& 1,427	& 1,526 \\
POS	& 16,678	& 13,123	& 1,709 & 1,846 \\
STS	& 3,073	& 2,073	& 500 & 500 \\
TC & 137,775 & 110,203 & 13,786 & 13,786 \\
QA & 14,239 & 11,255 & 1,492 & 1,429 \\
\hline
\end{tabular}
\caption{\label{tab:dataset-splits} Dataset splits with number of examples. }
\end{table}

\section{BERTa: A Model for Catalan}
\label{sec:modelcat}


Following other language-specific models, we train a RoBERTa \cite{DBLP:journals/corr/abs-1907-11692} base model (110M parameters), omitting the auxiliary Next Sentence Prediction task used in the original BERT, and just using the masked language modeling as the pre-training objective. The model is trained for 48 hours using 16 NVIDIA V100 GPUs of 16GB DDRAM, instead of 32 GB as in most works. For fitting an effective batch size of 2,048 sequences, we use gradient accumulation as implemented in Fairseq \cite{DBLP:journals/corr/abs-1904-01038}. Otherwise, we use the same hyperparameters as in the original RoBERTa, with a peak learning rate of 0.0005. We feed entire documents\footnote{We truncate those documents exceeding the maximum number of tokens.} instead of stand-alone sentences, fostering the modeling of long-range dependencies. We refer to the resulting model as BERTa. 

\section{Evaluation}
\label{sec:evaluation}

We evaluate BERTa comparing the performances with two well-known multilingual baselines, mBERT and XLM-RoBERTa,\footnote{In this work, we always use a base model version with 12 layers, for a fair comparison.} and another monolingual baseline, WikiBERT-ca (the Catalan WikiBERT by \citealp{pyysalo2020wikibert}), on a variety of downstream tasks.

\begin{table*}
\centering
\scalebox{0.9}{
\begin{tabular}{lllllll}
\hline
\textbf{model} & \textbf{NERC} & \textbf{POS} & \textbf{STS} & \textbf{TC} & \textbf{QA (ViquiQuAD)} & \textbf{QA (XQuAD)} \\
\hline
BERTa & 88.13 (2) & \textbf{98.97} (10) & 79.73 (5) & \textbf{74.16}  (9) & \textbf{86.97/72.29}  (9) & \textbf{68.89/48.87}  (9) \\
\hspace{3mm} + decontaminate      & \textbf{89.10} (6) & 98.94 (6) & \textbf{81.13} (8) & 73.84  (10) & 86.50/70.82  (6) & 68.61/47.26  (6) \\
 \hline
mBERT            & 86.38 (9) & 98.82 (9) & 76.34 (9) & 70.56 (10) & 86.97/72.22  (8) & 67.15/46.51  (8) \\
WikiBERT-ca        & 77.66 (9) & 97.60 (6) & 77.18 (10) & 73.22 (10) & 85.45/70.75 (10) & 65.21/36.60 (10) \\
XLM-RoBERTa & 87.66 (8) & 98.89 (10) & 75.40 (10)  & 71.68 (10) & 85.50/70.47  (5) & 67.10/46.42  (5) \\
\hline
\end{tabular}} 
\caption{\label{tab:results-tasks}
Results for the downstream tasks using different metrics. We use F1 for POS and NERC, accuracy for TC, an average of Pearson and Spearman coefficient for STS and F1/Exact Match for QA. We also report within round brackets the best epoch on the dev set.
}
\end{table*}

\subsection{Fine-tuning}
For evaluating our model against the existing baselines, we use common practices in the literature. For doing so, we leverage the Huggingface Transformers library \cite{DBLP:journals/corr/abs-1910-03771}. For each task, we attach a linear layer to the models and fine tune with the training set of the specific dataset. For tasks involving tokens classification, we use the first token of the last output layer. Specifically, in the case of the BERT model, we use the \texttt{[CLS]} token while for the RoBERTa models we use the \texttt{<s>} token. We train each model under the same settings (see Table \ref{tab:dataset-splits} for dataset splits) across tasks consisting of 10 training epochs, with an effective batch size of 32 instances, a max input length of 384 tokens and a learning rate of $5e^{-5}$. The rest of the hyperparameters are set to the default values in Huggingface Transformers.\footnote{ \url{https://github.com/huggingface/transformers/blob/master/src/transformers/training_args.py}} We select the best checkpoint as per the task-specific metric in the corresponding validation set, and then evaluate it on the test set. We report the results and metrics used in Table \ref{tab:results-tasks}. 

\subsection{Impact of fine-tuning data size}
\begin{figure}
    \centering
    \includegraphics[scale=0.47]{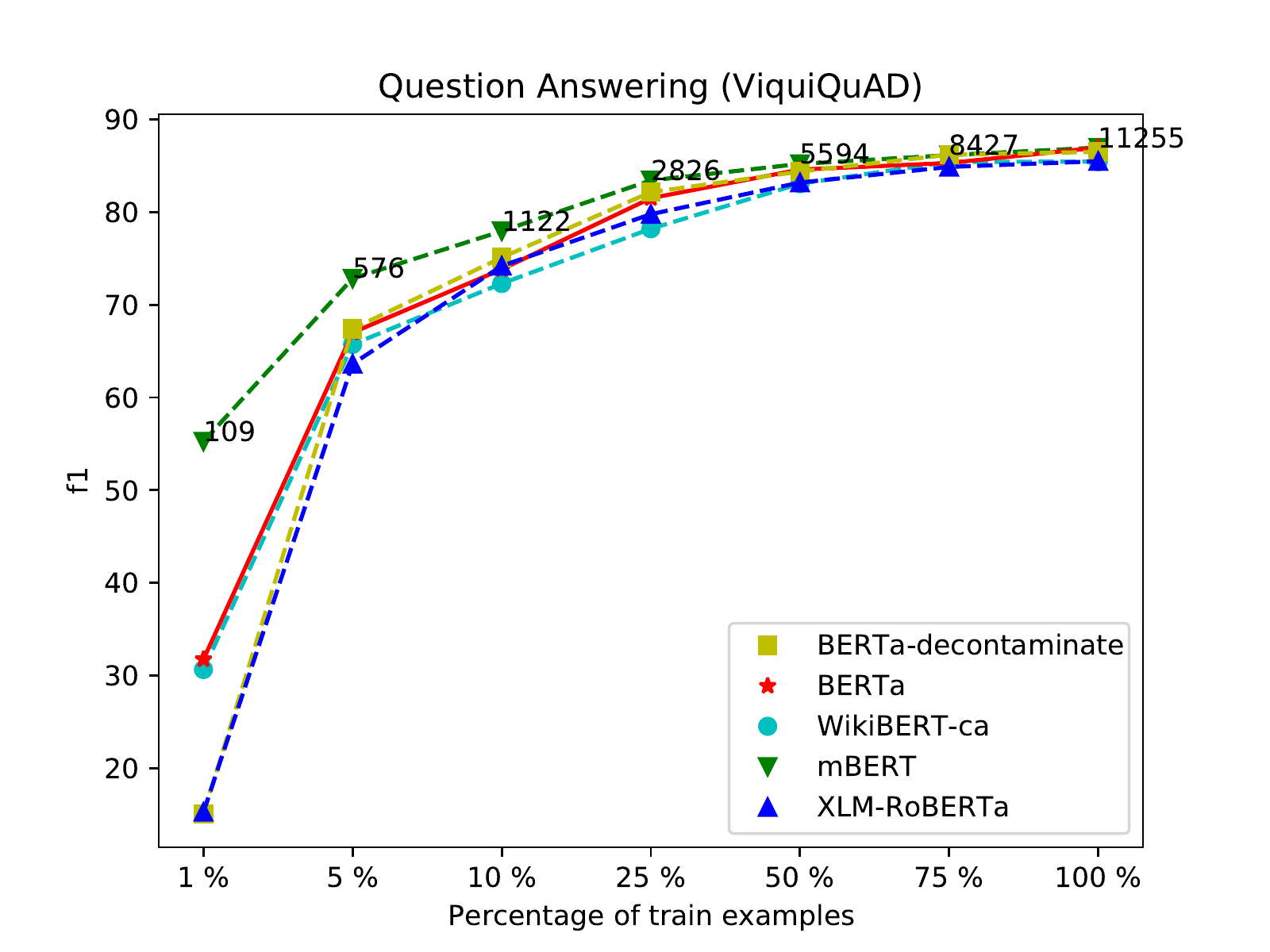}
    \includegraphics[scale=0.47]{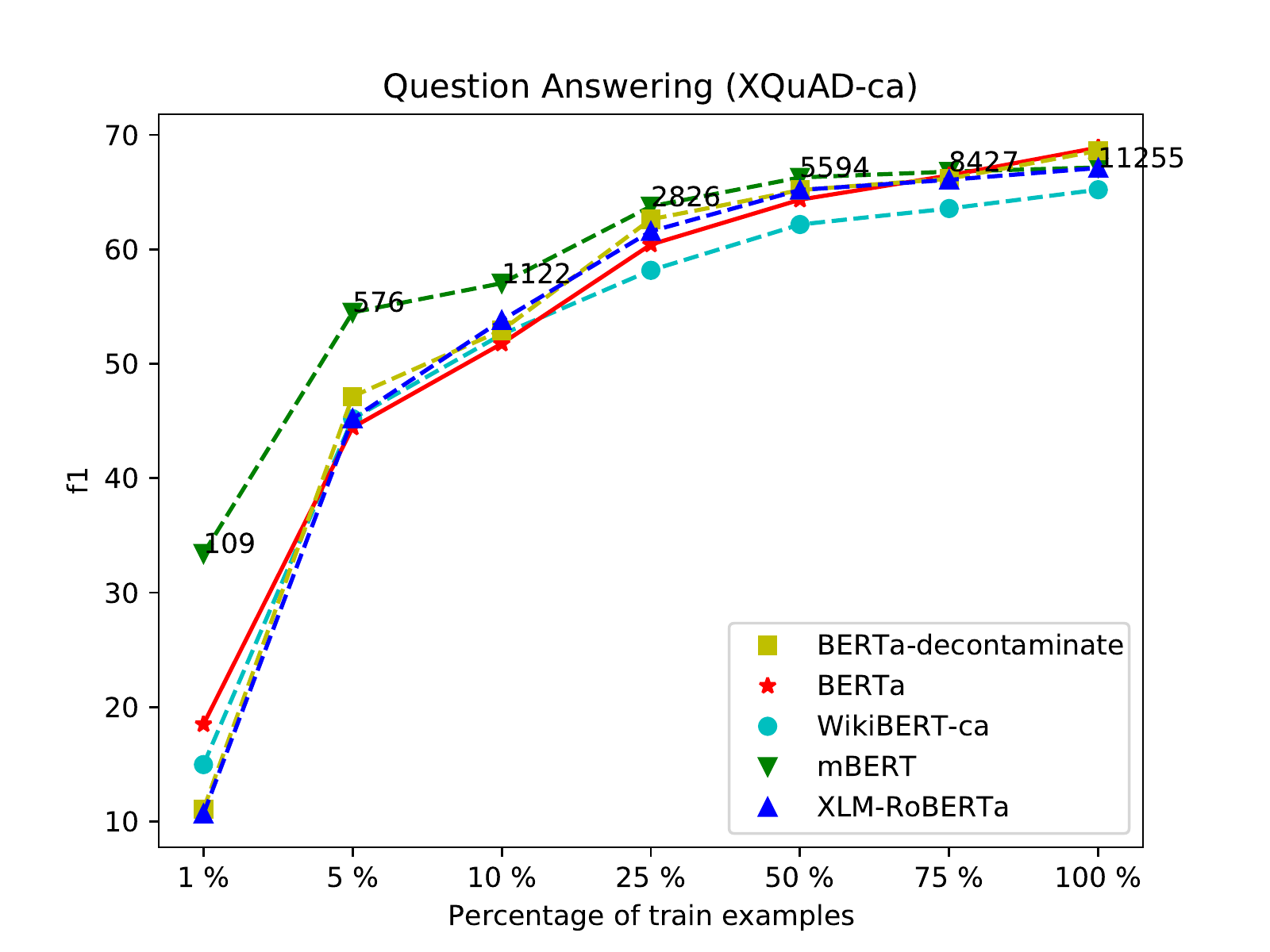}
    \caption{QA performance depending on the number of training examples.}
    \label{fig:qa_performance}
\end{figure}

\begin{figure}
    \centering
    \includegraphics[scale=0.45]{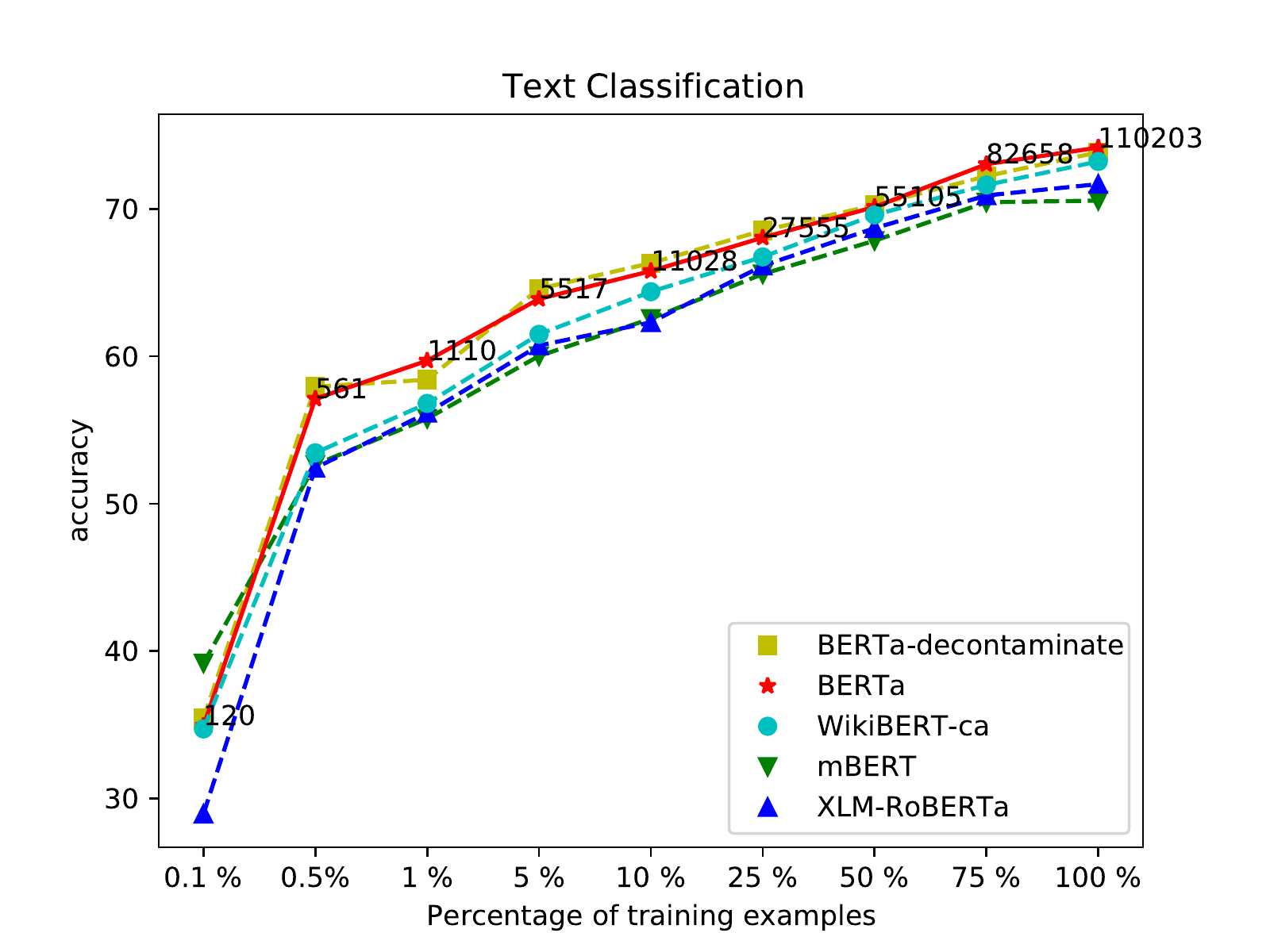}
    \caption{TC performance depending on the number of training examples.}
    \label{fig:tc_performance}
\end{figure}

To show the impact of the fine-tuning data size on the tasks performances, we incrementally increase the fine-tuning data size and fine-tune on downstream tasks. We choose TC and QA because they have enough data to study the size effect across several magnitudes ranging from a minimum of $10^2$ to a maximum of $10^4$ examples. In the case of TC, we reduce the fine-tuning data by performing a stratified sampling that preserves the original distribution of examples per label, while for QA we just sample random examples corresponding to a given size. Figures \ref{fig:qa_performance}, for QA, and \ref{fig:tc_performance}, for TC, show the results of the studied models in the test set of the corresponding tasks, when progressively increasing the amount of instances in the train set.

\subsection{Data contamination}
\label{sec:contamination}
As studied in \citet{brown2020language}, data contamination refers to the inclusion of content from test or development sets into the training corpora used for language model pre-training. Consequently, the effect of contamination might have an impact on the downstream tasks performances, distorting them. In our case, we measure the effect of data contamination produced by those sentences used during language model training and, after being annotated, as part of our evaluation benchmark. Specifically, the STS, TC, and QA benchmarks (unlike the NERC and POS ones) are indeed contaminated since some examples are built from the Catalan Wikipedia and ACN sources. Actually, about the 0.078\% of the training corpora (57,187 out of 73,172,152 sentences) are responsible for contamination. Therefore we train our BERTa model in two settings, one which includes these sentences and the other which does not include them. Note that all the compared models (mBERT, WikiBERT-ca, XLM-RoBERTa) have some degree of data contamination, since all of them include the Catalan Wikipedia as part of their training corpora. We show the results of BERTa along with BERTa-decontaminate in Table \ref{tab:results-tasks}. 



\section{Discussion}
\paragraph{Downstream tasks:}
Our model obtains better results in different benchmarking settings, ranging from simpler tasks with large training data sets (TC) to more advanced tasks with medium data (QA) and really complex tasks with small training data (STS).

In the case of NERC, BERTa clearly outperforms the existing baselines. In the case of POS, it also obtains better results, but only by a slight margin, being a considerably easy task, for which the existing models are already extremely competitive. For TC, BERTa also outperforms mBERT and XLM-RoBERTa models (by a larger margin when it has not been decontaminated). For QA, as measured in the ViquiQuAD dataset, only the  version of BERTa that has not been decontaminated equals mBERT and is slightly better than XLM-RoBERTa. Note, however, that both mBERT and XLM-RoBERTa are equally \textit{contaminated}, as pointed out in \ref{sec:contamination}. When measured in the  XQuAD-ca dataset, BERTa outperforms both models. Finally, for STS, BERTa outperforms mBERT and XLM-RoBERTa in 4.79 and 3.95 points respectively.  

Interestingly, in the case of the XQuAD test, we are evaluating with a test set from a different distribution (i.e. it does not belong to our original split). Here, BERTa also outperforms all the baselines.

\paragraph{Error analysis and Question Answering:}
For the QA task, many of the errors detected are due to the inclusion or not of articles and punctuation in the answers. Based on this observation, we adopt the strategy introduced in \citet{lewis-etal-2020-mlqa} and we ignore initial articles when evaluating. The F1 results improve by at least one percentage point, while the exact match metric increases even more when measured using this  MLQA\footnote{\url{https://github.com/facebookresearch/MLQA}} evaluation standard that discounts for initial articles. 

\paragraph{Size matters, quality too:}
One of the goals of our work has been to estimate the effect of corpus size in the results. Although to the best of our knowledge, CaText is one of smallest datasets used to train monolingual language models, we note that this data is roughly 8 times the size of the Catalan Wikipedia, used to train WikiBERT-ca, as well as the Catalan portion of mBERT. Interestingly, XLM-RoBERTa, built on a filtered version of CommonCrawl, contains a portion of Catalan comparable in size to CaText.\footnote{See Appendix in \citet{DBLP:journals/corr/abs-1911-02116}.}
In the line of FinBERT \cite{DBLP:journals/corr/abs-1912-07076}, we think that BERTa compensates for the relative scarcity of pre-training data with the quality of this data by using a rigorous filtering process.
This is supported by the fact that XLM-RoBERTa, with a Catalan portion comparable in size to our corpus, has a consistently lower performance, not only with respect to our model, but also with respect to mBERT, which contains a smaller but cleaner portion of Catalan.
While results for WikiBERT-ca seem to indicate that a Catalan model solely trained on the Wikipedia may not be enough in most cases, our results show that a model with sufficient diverse monolingual data that has been curated and cleaned, can outperform large multilingual models. Still, the question of how much data is enough remains unanswered.

\paragraph{Decontamination:}

We experimentally observe that our model generally benefits from data contamination in terms of performance, with a remarkable difference in the exact match metrics for QA, as shown in Table \ref{tab:results-tasks}.

A more intricate error analysis on the predictions for QA of both our BERTa models (with and without contamination) sheds light on the issue of decontamination. While the sole difference between these two is the inclusion of these sentences, we observe an improvement in exact match correct answers for the model that has not been decontaminated, and suspect this might be due to the fact that the model has indeed memorized the questions; nonetheless, more investigation is needed in this regard.

This posits the question of whether the metrics obtained in these kinds of benchmarks are only indicative of the actual performance in the tasks with sentences different from those in the pre-training corpus, or otherwise reflect a certain memorization of the evaluation sentences. Still, the decontaminated BERTa outperforms the baselines in all test sets (including the explicitly decontaminated ones) except for ViquiQuAD (in which mBERT is superior to the decontaminated BERTa), hinting that BERTa is better than the other models regardless of the possible effects of their contamination.

Note also that for the TC setting, the monolingual WikiBERT-ca model also beats multilingual models. In this scenario, neither WikiBERT-ca nor the multilingual models have unlabeled data from the training set in their pre-training datasets. It is, therefore, an uncontaminated scenario where a small monolingual model clearly beats the multilingual ones.
   
\paragraph{Varying fine-tuning data size in downstream tasks:}
We experiment with different data sizes for the train sets of the QA and TC tasks, as shown in Figure \ref{fig:qa_performance}.\footnote{This figure shows F1 score. In the Appendix \ref{app:qa}, we provide the results for the exact match evaluation as well.} Surprisingly, mBERT starts from a relatively high score, especially for QA, showing remarkably transfer capabilities when only a few examples are available. We point out that in the case of TC, as shown in Figure \ref{fig:tc_performance}, starting from about 500 examples, BERTa models always remain above the baselines. Overall, all the curves show a constant progression up to 75\% of training data and then seems to exhibit a slightly decreasing tendency in the case of multilingual models indicating they are approaching a plateau. Instead, monolingual ones still displays signs of improvement. For TC, Catalan-specific models, especially BERTa, show a better performance. Having a language-specific vocabulary may be helpful to further exploit increasing fine-tuning data.

\section{Conclusions}


We have demonstrated that it is worth to build a monolingual model for a moderately under-resourced language, even if it belongs to an over-represented linguistic family in multilingual models, such as the Romance language family. Our model outperforms the multilingual SOTA scores in all downstream tasks, as well as the monolingual scores from WikiBERT-ca, trained on a smaller and less varied corpus. Furthermore, by developing this model, we have contributed to the creation of open-source resources for Catalan, both for training and evaluation, that will encourage the development of technology in this language. We believe our methods can serve both as a recipe and a motivation for other languages in similar situations. In addition, we also release the cleaning pipeline used to build CaText, supporting 100+ languages.

As future work, we suggest further investigating the effect of contamination by explicitly studying the relation between memorization and performance in the exact match evaluation in QA. In addition, we propose conducting experiments for quantifying the effect of the data size, data cleaning, and data diversity (e.g., just Wikipedia vs. crawlings).

\section*{Acknowledgements}

This work was partially funded by the Generalitat de Catalunya through the project PDAD14/20/00001, the State Secretariat for Digitalization and Artificial Intelligence (SEDIA) within the framework of the Plan TL,\footnote{\url{https://www.plantl.gob.es/}} the MT4All CEF project,\footnote{\url{https://ec.europa.eu/inea/en/connecting-europe-facility/cef-telecom/2019-eu-ia-0031}} and the Future of Computing Center, a Barcelona Supercomputing Center and IBM initiative (2020).

We thank all the reviewers for their valuable comments.

\section*{Broader impacts}

Regarding possible environmental concerns, training new language-specific models is costly\footnote{See \url{https://www.bsc.es/marenostrum/marenostrum} for the estimated power consumption.} and one could argue that it could be avoided if a multilingual baseline performs well enough. However, we observe improvements that justify training from scratch another model. In addition, language-specific models are potentially more efficient, including inference, since their tokenizers generate less tokens.

As far as the model itself is concerned, we hypothesize that the pre-training corpus will have different biases and the model might reproduce them, so users must be aware of this issue. Nevertheless, with this work we contribute to an under-resourced language and open the door to follow similar approaches to other languages in similar situations, which can encourage a less English-centric view in the field of NLP.

\bibliographystyle{acl_natbib}
\bibliography{acl2021}

\clearpage
\appendix
\onecolumn

\section{Vocabularies}
\label{app:vocab}


\FloatBarrier
\begin{table}[h]
\centering
\begin{tabular}{cccc||r} 
\hline
BERTa & mBERT & WikiBERT-ca & XLM-RoBERTa & Tokens number \\
\hline
\xmark & & & & 52,000\\
 & \xmark& & & 119,547 \\
& & \xmark & & 20,101\\ 
& &  & \xmark & 250,002 \\
\hline
$\cap$ & $\cap$ & & & 11,251\\
$\cap$ & & $\cap$ & & 17,207\\ 
$\cap$ & & & $\cap$ & 13,063\\
\hline
\end{tabular}
\caption{\label{tab:tokens_intersect}
Number of tokens and vocabularies intersection.
}
\end{table}
\FloatBarrier

\section{Datasets}
\label{app:datasets}
\begin{figure}[!htb]
\centering
    \includegraphics[scale=0.55]{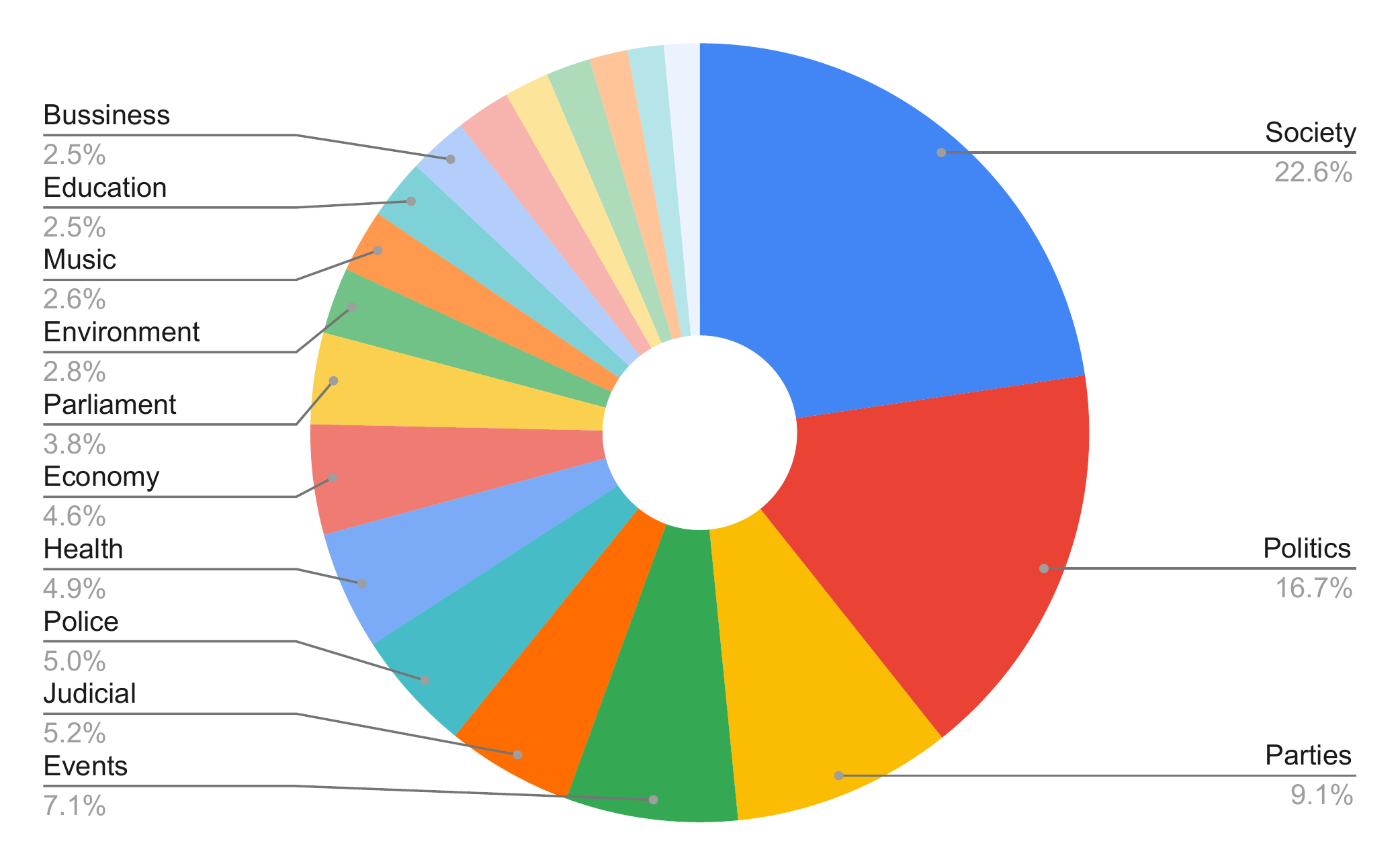}

    \caption{Label distribution of the Text Classification dataset. Here, we show the distribution of the filtered dataset, that is, keeping the labels with at least 2,000 instances.}
    \label{fig:text_classification_dataset}
\end{figure}

\begin{table}
\centering
\begin{tabular}{lrr} 
\hline
\textbf{ } &  \textbf{XQuAD-ca} & \textbf{ViquiQuAD}  \\
\hline
Paragraph & 48 & 597 \\
Context & 240 & 3,124 \\
Total sentences & 1,176 & 10,315 \\
Sentences/context & 4.9 & 3.30 \\
Tokens in context & 40,056 & 470,932 \\
\hline
Tokens in questions & 15,417 & 145,827 \\
Tokens in questions/questions & 12.96 & 9.59 \\
Tokens in questions/tokens in context & 0.38 & 0.31 \\
\hline
Tokens in answers & 4,436 & 63,596 \\
Tokens in answers/answers & 3.73 & 4.18 \\
Tokens in answers/tokens in context & 0.11 & 0.13 \\
  
\hline
\end{tabular}
\caption{\label{tab:tokens}
Statistics on the number of tokens in contexts, questions, and answers in our QA datasets
}
\end{table}

\begin{table}
\centering
\begin{tabular}{lrrrr} 
\hline
\textbf{ } &  \textbf{XQuAD-ca} & \textbf{ViquiQuAD} & \textbf{SQuAD} & \textbf{FQuAD} \\
\hline

   Lexical variation  &   33.0\%    & 7.0\%    & 33.3\%   &    35.2\% \\
   World knowledge     &   16.0\%    & 17.0\%    &  9.1\% &    11.1\%    \\
   Syntactic variation  &   35.0\%    & 43.0\%    &  64.1\%  &   57.4\%  \\
    Multiple sentence   &  17.0\%  & 9.0\%     & 13.6\%   &    17.6\%   \\
       
\hline
\end{tabular}
\caption{\label{tab:reasoning}
Question-answer reasoning typology. For XQuAD and ViquiQuad, we sampled 100 random question-answer pairs and classified them manually.
}
\end{table}



%

\begin{table*}
\centering
\begin{tabular}{lrrrr} 
\hline
\textbf{ } &  \textbf{XQuAD-ca} & \textbf{ViquiQuAD} & \textbf{XQuAD-en} & \textbf{FQuAD} \\
\hline

        Which? &   43.21\%    & 22.08\%    & 7.06\%   &    47.8\%  \\ 
             What? &   16.42\%    & 26.45\%    &  57.31\% &    4.1\%   \\ 
               Who? &   10.94\%    & 13.61\%    &  10.00\%  &   12.2\%   \\ 
            How many? &   8.3\%  & 5.72\%     & 6.55\%   &    5.6\%   \\ 
               How? &   8.02\%      & 12.41\%   & 5.13\%   &    6.8\%   \\ 
             When?   &   6.79\%   & 6.75\%     & 7.14\%   &    7.6\%   \\ 
             Why?  &   1.51\%  & 2.46\%      & 1.26\%   &     5.3\%   \\ 
           Where? &   3.58\%   & 10.3 \%    &  3.86\%  &   9.6\%    \\ 
        Other       & 1.23\%   & 0.16\%       &   1.93\%  &   1.00\%   \\
        
\hline
\end{tabular}
\label{tab:question-type}
\caption{
Question type frequencies. Differences between XQuAD-ca and XQuAD-en are explained because there is not an unique translation of the pronouns. }
\end{table*}
\clearpage
\section{Question Answering Evaluation}
\label{app:qa}
\begin{figure}[!htb]
    \centering
     \includegraphics[scale=0.75]{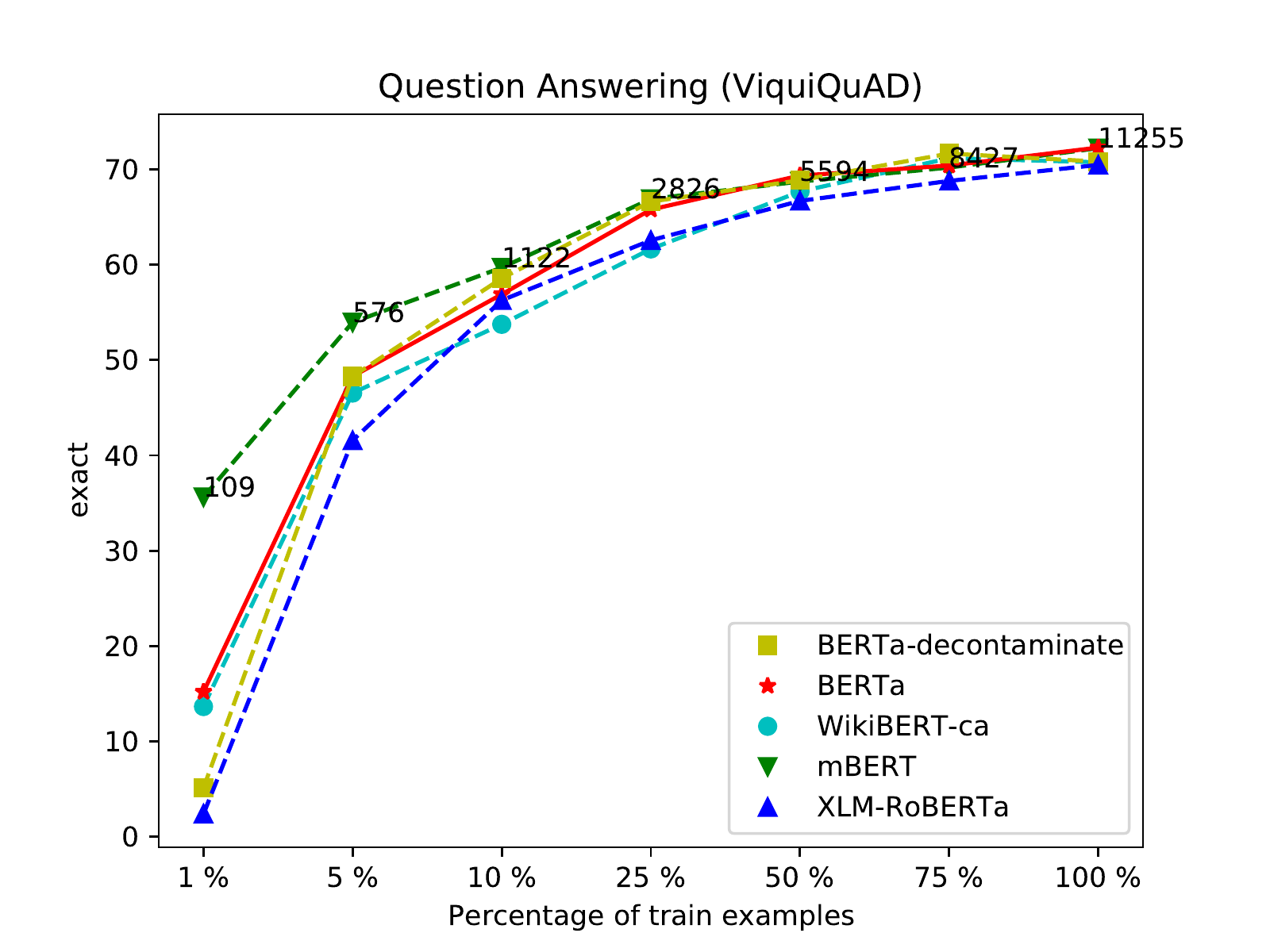}
 \includegraphics[scale=0.75]{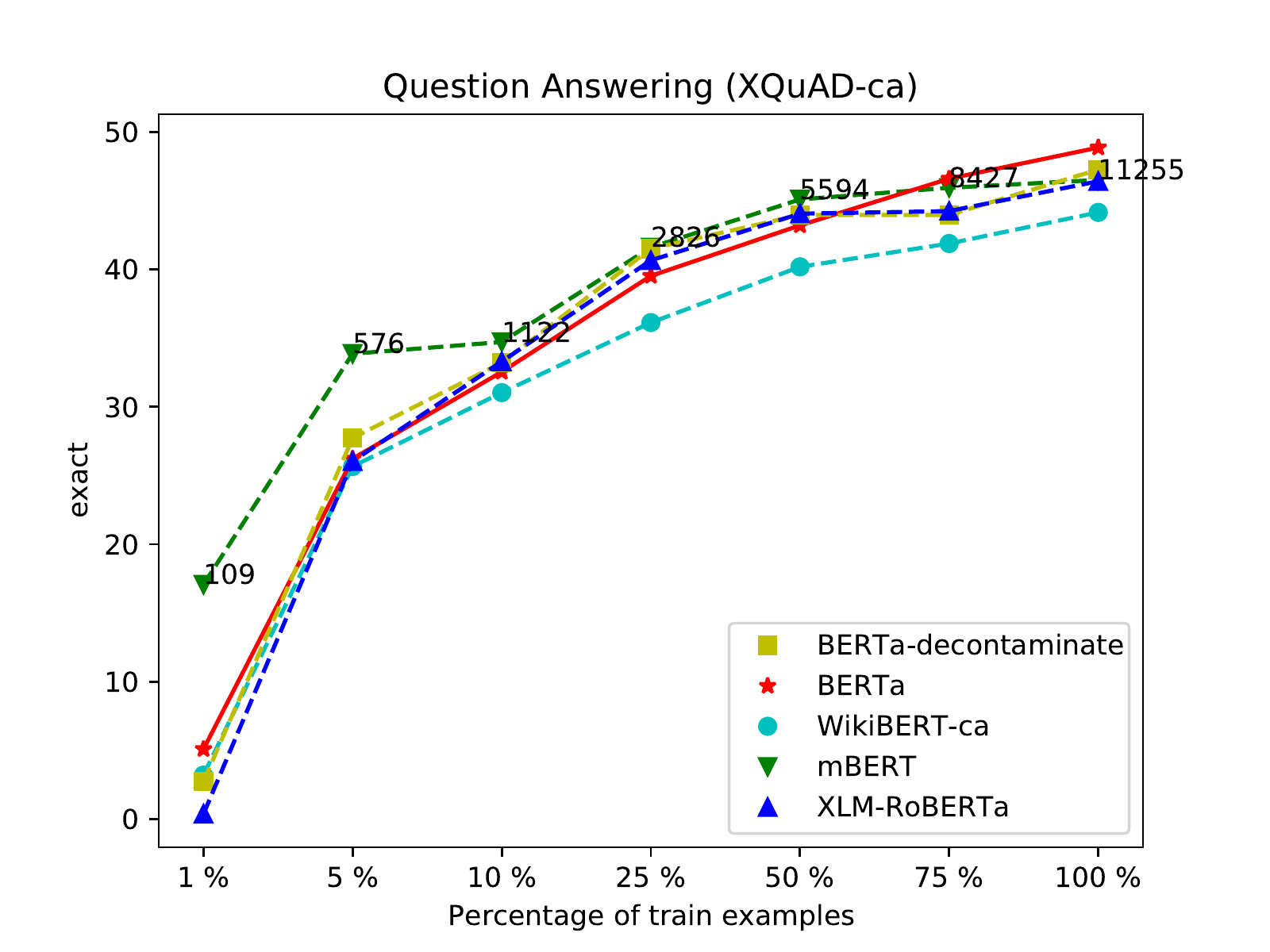}
    \caption{QA performance depending on the number of train instances (Exact match).}
    \label{fig:qa_performance_exact_match}
\end{figure}

\end{document}